\documentclass[]{fairmeta}
\usepackage{amsmath}
\usepackage{amsthm}
\usepackage{amssymb}
\usepackage{mathtools}  

\usepackage[most]{tcolorbox}
\usepackage{tabularx}
\usepackage{xcolor}
\usepackage{graphicx}
\usepackage{booktabs}
\usepackage{caption}

\usepackage{wrapfig}
\usepackage{graphicx}

\usepackage{xcolor}
\usepackage{mdframed}
\usepackage{tabularx, booktabs}

\definecolor{lightgray}{gray}{0.95}
\definecolor{lightblue}{RGB}{220,235,255}
\definecolor{midblue}{RGB}{60,90,160}
\definecolor{tealbar}{HTML}{56C4C0}
\definecolor{violetbar}{HTML}{A245EB}

\newmdenv[
  backgroundcolor=lightgray,
  linecolor=midblue,
  linewidth=1pt,
  skipabove=8pt,
  skipbelow=8pt,
  roundcorner=3pt,
  innertopmargin=6pt,
  innerbottommargin=6pt,
  innerleftmargin=8pt,
  innerrightmargin=8pt
]{graybox}

\newmdenv[
  backgroundcolor=lightblue,
  linecolor=midblue,
  linewidth=0.8pt,
  skipabove=6pt,
  skipbelow=6pt,
  roundcorner=2pt,
  innertopmargin=6pt,
  innerbottommargin=6pt,
  innerleftmargin=8pt,
  innerrightmargin=8pt,
]{bluebox}

\usepackage[ruled,vlined]{algorithm2e}

\usepackage{graphicx}
\usepackage[utf8]{inputenc} 
\usepackage[T1]{fontenc}    
\usepackage{url}            
\usepackage{booktabs}       
\usepackage{amsfonts}       
\usepackage{nicefrac}       
\usepackage{microtype}      
\usepackage{xcolor}         
\usepackage{multirow}
\usepackage{graphicx}
\usepackage{pifont}
\usepackage{makecell}

\newcommand{\fref}[1]{Figure~\ref{#1}}
\newcommand{\tref}[1]{Table~\ref{#1}}
\newcommand{\sref}[1]{\S\ref{#1}}

\usepackage{microtype}
\usepackage{graphicx}
\usepackage{amsmath}
\usepackage{bbm}
\usepackage{tcolorbox}

\usepackage{enumitem}
\usepackage{CJKutf8}

\usepackage[]{todonotes}

\newcounter{obssection}

\title{Multi-Task Reinforcement Learning for Enhanced Multimodal LLM-as-a-Judge}

\author[1,2,*]{Junjie Wu}
\author[1]{Xuan Kan}
\author[1]{Zihao He}
\author[1]{Shunwen Tan}
\author[1,3]{Bo Pan}
\author[1,\dagger]{Kaitai Zhang}

\affiliation[1]{Meta AI}
\affiliation[2]{Hong Kong University of Science and Technology}
\affiliation[3]{Emory University}

\contribution[*]{Work done during an internship at Meta}
\contribution[\dagger]{Corresponding Author}

\abstract{
Multimodal Large Language Models (MLLMs) have been widely adopted as MLLM-as-a-Judges due to their strong alignment with human judgment across various visual tasks. However, most existing judge models are optimized for single-task scenarios and struggle to generalize to diverse contexts, which is a critical requirement for reliable evaluation. To address this limitation, we propose Multi-Task Reinforcement Learning for MLLM-as-a-Judge (\textbf{MT-RL-Judge}), a framework that jointly optimizes the judge model across multiple tasks, leveraging the generalization capabilities of RL. Experimental results against several strong baselines demonstrate that MT-RL-Judge outperforms strong baselines in both judgment consistency and correlation with human preferences. Furthermore, our approach exhibits robust generalization on out-of-distribution tasks, further validating its effectiveness.
}

\date{\today}
\correspondence{Kaitai Zhang at \email{kaitaizhang@meta.com}}

\begin{document}

\maketitle

\section{Introduction}

The advancement of Multi-modal Large Language Models (MLLMs) has led to a proliferation of synthetic visual content. In industrial applications—ranging from intelligent customer service to advertisement generation—ensuring the quality and safety of these generated multimodal outputs has thus become a paramount task. However, evaluating such content remains a significant bottleneck~\citep{liu2023g,fu2024mme}. While human evaluation offers reliability, it is prohibitively expensive and difficult to scale to production-level processes. To address this challenge, the paradigm of \textit{MLLM-as-a-Judge}, which employs MLLMs as automated evaluators, has been proposed for large-scale evaluation~\citep{chen2024mllm,wang2025mllm,pu2025judge}.

Despite their promise, current MLLM-as-a-Judge frameworks encounter significant challenges in real-world production environments. First, relying solely on prompt engineering with off-the-shelf MLLMs often yields suboptimal performance, necessitating task-specific training to achieve high-quality judgments~\citep{chen2024mllm,zhou2025confprobench,luera2025mllm}. Second, existing trainable judges are typically specialized for narrow domains—such as safety compliance or image quality assessment—limiting their generalization to diverse evaluation scenarios~\citep{wang2025mllm,gu2025unime}. Furthermore, judges trained via Supervised Fine-Tuning (SFT) are prone to overfitting specific instruction formats. As evidenced in~\tref{tab:main_results}, an SFT model trained on pointwise alignment (single image-text pair) struggles to generalize to pairwise comparisons, rendering it brittle for dynamic industrial applications where requirements frequently evolve.

To address these scalability and robustness gaps, we propose a unified, Reinforcement Learning (RL)-enhanced framework: Multi-Task Reinforcement Learning for MLLM-as-a-Judge (\textbf{MT-RL-Judge}). Specifically, MT-RL-Judge leverages multi-task RL to train a comprehensive model capable of simultaneously handling diverse tasks with varying input-output formats. Unlike SFT-based judges, which tend to memorize superficial mappings between inputs and labels, our approach utilizes Group Relative Policy Optimization (GRPO)~\citep{shao2024deepseekmath} to incentivize the model to internalize the underlying evaluation logic. By explicitly generating reasoning steps prior to the final verdict, MT-RL-Judge significantly enhances both judgment quality and explainability. 

More importantly, compared to previous MLLM-as-Judges, MT-RL-Judge offers significant advantages for industrial deployment:
\begin{itemize}
    \item \textbf{Efficiency:} By unifying diverse evaluation tasks into a single judge model, we eliminate the need to switch between multiple specialized models when handling large-scale, heterogeneous inputs. This unification streamlines the inference pipeline and significantly reduces deployment costs.
    
    \item \textbf{Effectiveness:} We demonstrate that MT-RL-Judge does not compromise performance compared to single-task specialists; on the contrary, joint training across diverse tasks fosters a deeper understanding of evaluation logic, yielding superior results~\tref{tab:main_results}. Given that judgment accuracy is pivotal in industrial pipelines, MT-RL-Judge significantly enhances the reliability of automated quality assurance.
    
    \item \textbf{Generalization:} Crucial for real-world adaptability, MT-RL-Judge exhibits strong generalization across a broader range of evaluation tasks~\tref{tab:mjbench_results}. When evaluated on MJ-Bench~\citep{chen2024mj}—a dataset comprising task formats unseen during training (e.g., pairwise comparison)—our model significantly outperforms SFT counterparts, validating its reliability in handling novel evaluation scenarios without the need for retraining.
\end{itemize}

To the best of our knowledge, this work represents the first attempt to establish a unified, RL-based MLLM-as-a-Judge framework capable of generalizing across diverse evaluation tasks. This contribution not only offers a robust solution to the current evaluation bottleneck but also highlights a critical research direction for scalable, automated quality assurance in industrial scenarios.

\section{Related Works}
When evaluating open-ended model outputs, traditional reference-based metrics like BLEU~\citep{papineni2002bleu}, ROUGE~\citep{lin2004rouge}, and BERTScore~\citep{zhang2019bertscore} often correlate weakly with human preferences, necessitating more semantic-aware evaluation methodologies. To address this limitation, the \textit{LLM-as-a-Judge} paradigm emerged, which prompts capable LLMs to directly evaluate model outputs based on task-specific rubrics~\citep{zheng2023judging,gu2024survey}. This paradigm has subsequently been extended to the multimodal domain, leveraging MLLMs to process diverse sensory inputs, a framework formally referred to as \textit{MLLM-as-a-Judge}~\citep{chen2024mllm}.

Specifically, existing MLLM-as-a-Judge frameworks can be categorized as follows:
(1) \textbf{Prompt-based Judges:} These approaches directly employ capable, off-the-shelf MLLMs without additional parameter updates. Evaluation guidance is injected solely through prompt engineering, incorporating techniques such as detailed rubrics, Chain-of-Thought (CoT) reasoning paths, and in-context demonstrations~\citep{zheng2023judging, liu2023g, luera2025mllm, wang2025mllm}.
(2) \textbf{Fine-tuned Judges:} However, prompt-based approaches often struggle with intricate tasks requiring extensive context or domain-specific knowledge that cannot be effectively inserted solely on the input prompt. To address this limitation, recent works propose fine-tuning MLLMs on specific evaluation datasets via SFT or RL, thereby aligning the model more closely with the judging task to ensure reliable results~\citep{ko2025flex,pi2025mr}. Nevertheless, these fine-tuned judges often suffer from limited generalizability to unseen scenarios. Furthermore, most existing methods operate as specialized, single-task judges rather than a unified framework, rendering them impractical for deployment in large-scale commercial systems due to high maintenance and inference costs.

\section{Our Method}  
\subsection{Problem Formulation}

\paragraph{MLLM-as-a-Judge.}
Formally, let $\mathcal{D} = \{(x_i, y_i)\}_{i=1}^N$ denote a labeled dataset with $N$ examples, where $x_i$ represents the multimodal input (e.g., an image or image-instruction pair) and $y_i$ denotes the corresponding human-annotated label. For each input $x_i$ in $\mathcal{D}$ and a specific prompt $p_i$, an MLLM-as-a-Judge model $\mathcal{M}$ will take these information as inputs and produce an evaluation $\hat{y}_i = \mathcal{M}(x_i; p_i)$, aiming to make it approximate the human annotations $y_i$ with high fidelity.  

\paragraph{Unified MLLM-as-a-Judge.}
Standard MLLMs lack inherent alignment with the role of an evaluator, frequently resulting in unreliable judgments on unfamiliar tasks or complex criteria~\citep{chen2024mllm,wang2025mllm}. To mitigate this, existing research employs prompt engineering or post-training strategies—including Supervised Fine-Tuning (SFT) and Reinforcement Learning (RL)—to enhance the evaluative capabilities of foundation models~\citep{wang2025mllm,pi2025mr,ko2025flex}. However, these approaches predominantly focus on single-task optimization. Consequently, while such specialized judges may excel within narrow domains, they incur high deployment overheads and struggle to generalize across diverse evaluation scenarios, thereby hindering scalable deployment in commercial settings.

To address the scalability and generalization limitations of single-task models, the Unified MLLM-as-a-Judge paradigm aggregates multiple evaluation datasets into a comprehensive collection, denoted as $\mathcal{D}_{\text{unified}} = \bigcup_{k=1}^K \mathcal{D}_k$. The model is then jointly optimized across these datasets simultaneously using the following objective function:

\begin{equation}
\begin{split}
    \mathcal{L}(\theta) = - \mathbb{E}_{(x_i, p_i y_i) \sim \mathcal{D}_{\text{unified}}} \Big[ \sum_{i=1}^{N} \log P_\theta(y_i | x, p, y{<i}) \Big]
\end{split}
\end{equation}

While unified training exposes the MLLM-as-a-Judge to a broader spectrum of tasks, relying exclusively on SFT introduces a critical limitation. The standard SFT objective inherently encourages the model to mimic surface-level statistical correlations between inputs and outputs, rather than internalizing the underlying reasoning logic necessary for reliable judgments. Therefore, the model's generalization capability remains constrained, often leading to overfitting on specific prompt templates encountered during training.

\paragraph{RL-based MLLM-as-a-Judge with Reasoning.}
To enhance the reliability and generalizability of the judge model, another research direction integrates RL into MLLM-as-a-Judge training via reward modeling. This approach specifically encourges the model to employ a ``reasoning before answering'' strategy during the evaluation process~\citep{pi2025mr}. By explicitly generating a reasoning trace prior to the final prediction, the judge model can more accurately approximate the internal evaluation logic aligned with human preferences, ultimately leading to superior judgment performance. 

\subsection{MT-RL-Judge}
\label{sec:our method}

While Unified SFT improves task coverage via data aggregation, it remains constrained by the inherent limitations of maximum likelihood estimation. Furthermore, existing RL-based judges are predominantly confined to isolated domains, leaving the potential of unified, multi-task RL evaluation largely unexplored. To bridge this gap, we propose \textbf{MT-RL-Judge}, a framework that optimizes a global policy to maximize the expected composite reward across diverse judging tasks simultaneously. This paradigm shifts the objective from merely fitting specific dataset distributions to including a generalized reasoning mechanism that is both robust and transferable across varying contexts.

\paragraph{Reward Function.}
Specifically, the reward function for training MT-RL-Judge is formulated as a weighted combination of two complementary components: the format reward and the accuracy reward. The format reward ($R_{\text{For}}$) ensures that the model output adheres to the requisite structure, specifically the ``reasoning-first'' paradigm. Conversely, the accuracy reward ($R_{\text{Acc}}$) enables the generation of reliable reasoning traces that culminate in correct judgments. Formally, these rewards are defined as:

\begin{equation}
    R_{\text{Acc}} = 
    \begin{cases} 
        1.0 & \text{if } \hat{y} = y \\
        0.0 & \text{otherwise}
    \end{cases}
\end{equation}

\begin{equation}
    R_{\text{For}} = 
    \begin{cases} 
        1.0 & \text{if the format is followed} \\
        0.0 & \text{otherwise}
    \end{cases}
\end{equation}

The total reward is then computed as a linear combination of the two rewards:
\begin{equation}
    R_{\text{total}} = (1-\alpha) \cdot R_{\text{Acc}} + \alpha \cdot R_{\text{For}}
\end{equation}
where $\alpha$ is the weighting hyperparameter that governs the relative importance of the two rewards.

\paragraph{Training Objective.}
To optimize the judge model, we employ Group Relative Policy Optimization (GRPO)~\citep{shao2024deepseekmath}, which eliminates the need for a separate value function by utilizing the average reward computed across a group of generated outputs. Formally, for each input prompt sampled from the unified dataset $\mathcal{D}_{\text{unified}}$, we generate a group of $G$ outputs conditioned on the same prompt by sampling from the judge model multiple times. The reward value at this step is then derived by averaging the reward values obtained from these $G$ generations.

Regarding the overall training objective, instead of maximizing rewards for a single isolated task, MT-RL-Judge seeks the optimal parameters that maximize the expected reward across the entire unified dataset, formulated as follows:

\begin{equation}
    \theta^* = \mathop{\arg\max}_{\theta} \mathbb{E}_{(x, p, y) \sim \mathcal{D}_{unified}} \left[ R_{\text{total}}(\mathcal{M}_\theta(x)) \right]
\end{equation}

\begin{table*}[htbp] 
\resizebox{\textwidth}{!}{%
\begin{tabular}{lcccccc}
\toprule
\textbf{Split} & \textbf{AGIN-Nat} & \textbf{AGIN-Tech} & \textbf{AGIN-Rat} & \textbf{Seetrue} & \textbf{Unsafe Bench} & \textbf{Image Reward} \\
\midrule
\textbf{Train} & 4,839 (2440/2399) & 4,839 (1422/3417) & 4,839 (1652/3187) & 5,544 (2535/3009) & 7,298 (2954/4344) & 6,194 (1690/4504) \\
\textbf{Val}   & 605 (285/320)     & 605 (156/449)     & 605 (183/422)     & 693 (302/391)     & 811 (317/494)     & 2,584 (968/1616)  \\
\textbf{Test}  & 605 (300/305)     & 605 (158/447)     & 605 (187/418)     & 693 (309/384)     & 2,037 (777/1260)  & 2,720 (588/2132)  \\
\bottomrule

\end{tabular}%
}
\caption{Statistics of the datasets used in our experiments. Values in parentheses denote the total number of (negative/positive) samples for each task across different data splits.}
\label{tab:dataset_statistics}
\end{table*}

Through this unified optimization process, MT-RL-Judge consistently delivers high-quality evaluations across a diverse range of tasks. Crucially, the explicit generation of high-quality reasoning traces renders these judgments highly interpretable. This combination of accuracy and transparency ensures that the model is both reliable for deployment in industrial applications and closely aligned with human preferences.

\section{Experiments}

\subsection{Datasets}

We evaluate our proposed framework on six benchmark datasets spanning three distinct capabilities: text-image alignment, safety compliance, and visual quality assessment. Specifically, we utilize \textbf{SeeTRUE}~\citep{yarom2023you} and \textbf{ImageReward}~\citep{xu2023imagereward} to assess the semantic consistency between images and text prompts. For safety evaluation, we employ \textbf{UnsafeBench}~\citep{qu2025unsafebench} to detect harmful visual content. Additionally, we incorporate three subsets from the \textbf{AGIN} benchmark~\citep{chen2023exploring}—Naturalness, Rationality, and Technical Quality—to scrutinize the perceptual quality of generated images. The detailed statistics for each dataset are summarized in Table~\ref{tab:dataset_statistics}.

\subsection{Setting}
To evaluate the effectiveness of our proposed MT-RL-Judge, we benchmark it against several MLLM-as-a-Judge baselines, all implemented using the same foundational backbone. First, we establish a zero-shot baseline using an Off-the-shelf MLLM, which evaluates inputs directly via instructional prompts without any task-specific fine-tuning. Next, we compare our method against SFT-based models, including single-task SFT judges trained exclusively on individual evaluation tasks (\textbf{SFT-Single}), and an unified SFT iudge trained on an aggregated dataset encompassing all tasks (\textbf{SFT-Unified}). Finally, to isolate the benefits of multi-task synergy within the RL paradigm, we include single-task RL judges (\textbf{RL-Single}), which apply the same RL reward modeling technique but are trained on each task independently. Throughout our experiments, we utilize \texttt{Qwen3-VL-30B-A3B-Instruct} as the backbone model. Detailed experimental configurations and specific prompt templates are provided in Appendix~\sref{app:training_config} and Appendix~\sref{app:prompt}, respectively. Since our evaluation tasks are mainly binary classification problems, we adopt the Macro-F1 score as our primary evaluation metric.

\subsection{Main Results}

\begin{table*}[htbp]
\centering
\label{tab:main_results_grouped_by_method}
\resizebox{140mm}{!}{%
\begin{tabular}{lcccccc}
\toprule
\textbf{Method} & \textbf{AGIN-Nat.} & \textbf{AGIN-Tech.} & \textbf{AGIN-Rat.} & \textbf{Seetrue} & \textbf{ImageReward} & \textbf{Unsafe Bench} \\
\midrule
Off-the-shelf & 67.99 & 63.24 & 64.77 & 80.01 & 55.07 & 72.78 \\
\midrule
SFT-Single   & 78.64 & 77.04 & 78.08 & 80.41 & 64.95 & \textbf{90.28} \\
SFT-Unified  & \textbf{81.75} & \underline{81.22} & 81.31 & 82.32 & 63.34 & \underline{89.49} \\
\midrule
RL-Single    & 80.50 & 80.77 & \textbf{82.71} & \underline{83.41} & \textbf{65.07} & 86.92 \\
MT-RL-Judge   & \underline{81.63} & \textbf{81.37} & \underline{81.58} & \textbf{83.67} & \underline{64.97} & 85.22 \\
\bottomrule
\end{tabular}%
}
\caption{Macro-F1 results on all the judging tasks, and the best performance on each task is highlighted in \textbf{bold}, while the second highest results is \underline{underlined}.}
\label{tab:main_results}
\end{table*}

\tref{tab:main_results} presents the comparative evaluation results. We highlight three key observations from the results:

\paragraph{RL Enhances MLLM-as-a-Judge.}
As shown, RL-based judges consistently outperform their SFT-based counterparts across the majority of evaluation tasks. For instance, RL-Single surpasses SFT-Single on 5 out of 6 benchmarks, achieving notable gains on SeeTrue (+3.0) and AGIN-Rationality (+4.63). These improvements are largely attributable to the reasoning-intensive nature of these specific tasks. This validates our hypothesis: whereas SFT tends to mimic surface-level statistical patterns, RL-based training actively incentivizes the MLLM-as-a-Judge to engage in rigorous logical deduction prior to rendering a final prediction, ultimately yielding more reliable evaluations.

\paragraph{Unified Training Enhances Generalization.}
Surprisingly, aggregating diverse evaluation tasks into a unified MLLM-as-a-Judge framework does not lead to significant performance degradation; rather, it frequently yields superior judging quality compared to isolated training. For example, the SFT-Unified judge outperforms SFT-Single on the majority of tasks (e.g., achieving 81.75 versus 78.64 on AGIN-Nat.). This suggests that multi-task exposure enables the judge model to capture shared evaluation criteria and latent correlations across different domains, thereby preventing it from overfitting to task-specific prompt instructions.

\paragraph{Effectiveness of MT-RL-Judge.}
By synthesizing the aforementioned strengths, MT-RL-Judge consistently achieves the best overall performance across diverse benchmarks (e.g., 83.67 on SeeTrue). Although SFT-Single scores marginally higher on UnsafeBench—likely due to its tendency to memorize dataset-specific safety patterns shared between the training and test splits—MT-RL-Judge maintains a highly competitive standard across all other evaluation tasks. Ultimately, this demonstrates that the synergy between unified multi-task exposure and RL-driven reasoning yields a remarkably robust and reliable evaluator.

\subsection{MT-RL-Judge Enhances Generalizability}

As demonstrated in~\tref{tab:main_results}, MT-RL-Judge exhibits strong generalization capabilities across diverse scenarios, an advantage inspiring by its deeper comprehension of the evaluation criteria facilitated by the explicit generation of reasoning traces. To rigorously investigate this out-of-domain generalization potential, we conduct an evaluation on MJ-Bench~\citep{chen2024mj}, a dataset strictly held out from the training corpora of all judge models evaluated in~\tref{tab:main_results}.

Specifically, while our judge models were trained exclusively on pointwise evaluation tasks (e.g., outputting a binary ``yes'' or ``no'' for a single image), MJ-Bench requires the model to perform pairwise comparisons (i.e., selecting the superior candidate from two images) for image-text matching and image safety assessments, where both task types appear in the training data of such judge models. This distinct setup evaluates the model's capability to resolve the same underlying task semantics under a novel input formulation. Consequently, it serves as a critical test to determine whether the judge model has genuinely internalized the intrinsic logic of judging, or if it has just memorized the specific prompt templates of the training distribution.

\begin{table}[htbp]
\centering
\begin{tabular}{lcccc}
\toprule
\textbf{Method} & \textbf{Image-text Alignment} & \textbf{Safety Judge} \\
\midrule
Off-the-shelf       & 59.41 & 73.07 \\
SFT-Unified     & 55.82          & 49.40 \\
MT-RL-Judge      & \textbf{60.59}          & \textbf{82.23} \\
\bottomrule
\end{tabular}
\caption{Evaluation results (Macro-F1) on MJ-Bench. The best performance per task is highlighted in \textbf{bold}.}
\label{tab:mjbench_results}
\end{table}

\paragraph{Results.}
The results on MJ-Bench are summarized in Table~\ref{tab:mjbench_results}, where we observe a significant contrast between the generalization capabilities of SFT-Unified and MT-RL-Judge:

\begin{itemize}
    \item \textbf{SFT Overfits to Specific Task Formats:} The SFT-Unified judge struggles significantly with the unseen pairwise format of MJ-Bench, despite being trained on tasks with identical underlying semantics. Notably, its performance on the Safety Judge task degrades to 49.40\%, falling substantially below that of the zero-shot Off-the-shelf baseline (73.07\%). This stark degradation indicates that even a unified SFT judge, despite its multi-task exposure, lacks robust generalization capabilities. Instead, it strongly overfits to the single-image input structure prevalent in its training corpus, failing to adapt when the visual context expands to encompass multiple candidate images.
    
    \item \textbf{RL Enables Robust Generalization:} In contrast, MT-RL-Judge demonstrates superior out-of-domain generalizability. It not only adapts seamlessly to the unseen pairwise evaluation format of MJ-Bench, but also delivers highly competitive performance, achieving \textbf{60.59\%} on the Alignment task and \textbf{82.23\%} on the Safety task. This validates our hypothesis that the RL-driven reasoning process encourages the model to abstract the fundamental evaluation criteria (e.g., the intrinsic definitions of safety and alignment). Consequently, the judge model is empowered to flexibly extrapolate these learned principles to novel task formulations completely absent from its training distribution.
\end{itemize}

\section{Conclusion}

In this paper, we propose \textbf{MT-RL-Judge}, a unified multi-task reinforcement learning framework designed to enhance MLLM-as-a-Judge evaluators. By jointly optimizing diverse evaluation tasks with a composite reward encompassing both structural format and prediction accuracy, our approach incentivizes the model to internalize the underlying judging logic rather than overfit to surface-level instruction formats. Comprehensive experiments across six distinct tasks demonstrate that MT-RL-Judge consistently outperforms various strong baselines. Crucially, MT-RL-Judge exhibits robust out-of-domain generalization on unseen pairwise formats, highlighting its resilience to distribution shifts. Our results suggest that unified multi-task RL judge represents a highly promising direction for scalable and reliable multimodal evaluation in industrial applications.

\section{Ethical Considerations}

While the proposed approach is targeted to be deployed in an industrial context, the experimental results presented in this manuscript are based solely on publicly available datasets. As such, this work does not introduce additional ethical considerations.

\bibliographystyle{assets/plainnat}
\bibliography{paper}

\clearpage
\newpage
\beginappendix
\section{Training Configurations}
\label{app:training_config}

For the SFT stage, we utilize the LLaMA-Factory framework~\citep{zheng2024llamafactory}, while for Reinforcement Learning (RL), we employ EasyR1~\citep{zheng2025easyr1}. All MLLM-as-a-Judge models in our experiments are initialized from the \texttt{Qwen/Qwen3-VL-30B-A3B-Instruct} base model.

Specifically, SFT is performed via full-parameter fine-tuning using the AdamW optimizer with a cosine learning rate schedule. To efficiently handle high-resolution visual inputs, we enable Flash Attention 2. Detailed hyperparameters are provided in Table~\ref{tab:sft_config}. We conduct training until performance on the validation set plateaus, subsequently selecting the checkpoint with the highest validation score for downstream experiments. For hyperparameters not explicitly listed, we adhere to the standard configuration defaults of LLaMA-Factory. 

\begin{table}[h]
    \centering
     \setlength{\tabcolsep}{1mm}
    \begin{tabular}{lc}
        \toprule
        \textbf{Hyperparameter} & \textbf{Value} \\
        \midrule
        Precision & bfloat16 \\
        Learning Rate & $1.0 \times 10^{-5}$ \\
        Weight Decay & $1.0 \times 10^{-5}$ \\
        Optimizer & AdamW \\
        Batch Size & 256 \\
        Max Image Resolution & 4,194,304 pixels \\
        \bottomrule
    \end{tabular}
    \caption{Hyperparameters for SFT.}
    \label{tab:sft_config}
\end{table}

As for the RL stage, we mainly employ the Group Relative Policy Optimization (GRPO) algorithm, and set the number of generations per prompt (rollout $N$) to 20 to ensure sufficient exploration. The specific configuration is detailed in Table~\ref{tab:rl_config}. We conduct training until reward on the validation set plateaus, subsequently selecting the checkpoint with the highest accuracy reward for downstream experiments. For hyperparameters not explicitly listed, we adhere to the standard configuration defaults of Easy-R1. 

\begin{table}[h]
    \centering
         \setlength{\tabcolsep}{1mm}
    \begin{tabular}{lc}
        \toprule
        \textbf{Hyperparameter} & \textbf{Value} \\
        \midrule
        Precision & bfloat16 \\
        Optimizer & AdamW \\
        Global Batch Size & 256 \\
        Rollout Batch Size & 512 \\
        Rollout ($N$) & 20 \\
                Max Image Resolution & 4,194,304 pixels \\
        \bottomrule
    \end{tabular}
        \caption{Hyperparameters for RL.}
    \label{tab:rl_config}
\end{table}

\section{Prompt for all the Tasks}
\label{app:prompt}
In this section, we detail the full set of prompts used in our experiments. Specifically, the SFT prompts for the six tasks listed in~\tref{tab:dataset_statistics} are presented in~\fref{fig:sft_prompt_unsafe},~\fref{fig:sft_prompt_agin_tech},~\fref{fig:sft_prompt_agin_rat},~\fref{fig:sft_prompt_agin_nat},~\fref{fig:sft_prompt_seetrue}, and~\fref{fig:sft_prompt_image_reward}, respectively. Correspondingly, the RL prompts for these tasks are shown in~\fref{fig:rl_prompt_unsafe},~\fref{fig:rl_prompt_agin_tech},~\fref{fig:rl_prompt_agin_rat},~\fref{fig:rl_prompt_agin_nat},~\fref{fig:rl_prompt_seetrue}, and~\fref{fig:rl_prompt_image_reward}.

\begin{figure*}
  \begin{tcolorbox}
 You are an experienced content moderation specialist with expertise in digital safety and content policy enforcement. Your task is to evaluate whether a given image contains safe or unsafe content. \\

You will be provided the following information:\\

Image Category: \{\} \\
Image Description: \{\} \\
Image Source: \{\} \\
The image to evaluate: <|image|> \\

Please carefully analyze the image and the provided information to determine if the image contains any unsafe content, including but not limited to: \\
- Violence or gore \\
- Adult or sexual content\\
- Hate symbols or discriminatory imagery\\
- Dangerous activities or self-harm\\
- Illegal substances or activities\\
- Harassment or bullying content\\
- Misinformation or manipulated media\\

Respond with a single word: "Safe" if the image does not contain unsafe content, or "Unsafe" if it contains any unsafe content. Do not provide any explanation.
    \end{tcolorbox}
    \caption{Prompt for SFT on Unsafe Bench.}
    \label{fig:sft_prompt_unsafe}
\end{figure*}

\begin{figure*}
  \begin{tcolorbox}
 You are an experienced image quality assessment expert with expertise in evaluating technical aspects of digital imagery. Your task is to evaluate whether a given image has good technical quality. \\

You will be provided the following information: \\

The image to evaluate: <|image|> \\

Please carefully analyze the image's technical quality based on the following criteria:\\
- Sharpness and clarity: Is the image clear and well-focused, or is it blurry and out of focus?\\
- Compression artifacts: Are there visible pixelation, blocky patterns, or compression-related distortions?\\
- Noise and grain: Does the image have excessive noise, grain, or visual static?\\
- Distortion: Are there any warping, stretching, or other geometric distortions?\\
- Resolution: Is the image resolution sufficient for clear viewing, or is it overly low-resolution?\\
- Overall visual integrity: Can the image content be clearly perceived without significant technical impediments?\\

Consider the image to have "good technical quality" if it has minimal to no technical issues that would impair viewing or comprehension of the content.\\

Respond with a single word: "Yes" if the image has good technical quality, or "No" if it does not have good technical quality. Do not provide any explanation.
    \end{tcolorbox}
    \caption{Prompt for SFT on AGIN-Tech.}
    \label{fig:sft_prompt_agin_tech}
\end{figure*}

\begin{figure*}
  \begin{tcolorbox}
 You are an experienced visual content analyst with expertise in evaluating the logical consistency and rationality of images. Your task is to evaluate whether a given image is rational. \\

You will be provided the following information: \\

The image to evaluate: <|image|> \\

Please carefully analyze the image's rationality based on the following criteria:\\
- Physical plausibility: Do objects, people, and scenes follow realistic physical laws and proportions?\\
- Logical consistency: Do the elements in the image make sense together in the given context?\\
- Anatomical accuracy: Are human or animal bodies depicted with correct anatomy and natural poses?\\
- Spatial coherence: Are objects positioned and sized appropriately relative to each other?\\
- Contextual appropriateness: Do the elements fit logically within the scene or setting?\\
- Absence of absurdities: Are there any impossible, nonsensical, or surreal elements that defy rationality?\\

Consider the image to be "rational" if it depicts a coherent, logically consistent scene with minimal to no irrational or absurd elements.\\

Respond with a single word: "Yes" if the image is rational, or "No" if it is not rational. Do not provide any explanation.
    \end{tcolorbox}
    \caption{Prompt for SFT on AGIN-Rat.}
    \label{fig:sft_prompt_agin_rat}
\end{figure*}

\begin{figure*}
  \begin{tcolorbox}
 You are an experienced image authentication expert with expertise in distinguishing AI-generated images from real photographs. Your task is to evaluate whether a given image looks natural. \\

You will be provided the following information: \\

The image to evaluate: <|image|> \\

Please carefully analyze the image's naturalness based on the following criteria:\\
- AI generation artifacts: Are there visible signs of AI generation such as unnatural textures, blending issues, or synthetic patterns?\\
- Photorealism: Does the image exhibit the natural characteristics of a real photograph, including authentic lighting, shadows, and depth?\\
- Detail consistency: Are fine details (skin texture, hair strands, fabric, surfaces) rendered naturally and consistently?\\
- Object and human features: Do faces, hands, objects, and their features appear natural without distortions or uncanny elements?\\
- Visual coherence: Does the overall image maintain photographic quality without obvious artificial or synthetic areas?\\

Consider the image to be "natural" if it looks as authentic as a real photograph with minimal to no visible unnatural or AI-generated characteristics.\\

Respond with a single word: "Yes" if the image looks natural, or "No" if it does not look natural. Do not provide any explanation.
    \end{tcolorbox}
    \caption{Prompt for SFT on AGIN-Nat.}
    \label{fig:sft_prompt_agin_nat}
\end{figure*}

\begin{figure*}
  \begin{tcolorbox}
 You are an experienced visual-linguistic analyst with expertise in evaluating the alignment between textual descriptions and visual content. Your task is to evaluate whether a given text description matches a given image.\\

You will be provided the following information: \\

Text Description: \{Image Description\}\\
The image to evaluate: <|image|> \\

Please carefully analyze whether the text description accurately matches the image based on the following criteria:\\
- Content accuracy: Do the objects, people, animals, or scenes described in the text appear in the image?\\
- Detail consistency: Are the specific details mentioned in the text (colors, positions, actions, attributes) consistent with what is shown in the image?\\
- Completeness: Does the text describe the main elements visible in the image without introducing elements that are not present?\\
- Contextual alignment: Does the overall context or scene described in the text match the visual context of the image?\\

Consider the text and image to "match" if the text accurately describes the visual content of the image without significant discrepancies or contradictions.\\

Respond with a single word: "Yes" if the text matches the image, or "No" if the text does not match the image. Do not provide any explanation.
    \end{tcolorbox}
    \caption{Prompt for SFT on SeeTrue.}
    \label{fig:sft_prompt_seetrue}
\end{figure*}

\begin{figure*}
  \begin{tcolorbox}
You are an experienced visual-linguistic analyst with expertise in evaluating the alignment between textual descriptions and visual content. Your task is to evaluate whether a given text description matches a given image.\\

You will be provided the following information: \\

Text Description: \{Image Description\}\\
The class of the image: \{Classification\}\\
The image to evaluate: <|image|> \\

Please carefully analyze whether the text description accurately matches the image based on the following criteria:\\
- Content accuracy: Do the objects, people, animals, or scenes described in the text appear in the image?\\
- Detail consistency: Are the specific details mentioned in the text (colors, positions, actions, attributes) consistent with what is shown in the image?\\
- Completeness: Does the text describe the main elements visible in the image without introducing elements that are not present?\\
- Contextual alignment: Does the overall context or scene described in the text match the visual context of the image?\\

Consider the text and image to "match" if the text accurately describes the visual content of the image without significant discrepancies or contradictions.\\

Respond with a single word: "Yes" if the text matches the image, or "No" if the text does not match the image. Do not provide any explanation.
    \end{tcolorbox}
    \caption{Prompt for SFT on Image Reward.}
    \label{fig:sft_prompt_image_reward}
\end{figure*}


\begin{figure*}
  \begin{tcolorbox}
 You are an experienced content moderation specialist with expertise in digital safety and content policy enforcement. Your task is to evaluate whether a given image contains safe or unsafe content. \\

You will be provided the following information:\\

Image Category: \{\} \\
Image Description: \{\} \\
Image Source: \{\} \\
The image to evaluate: <|image|> \\

Please carefully analyze the image and the provided information to determine if the image contains any unsafe content, including but not limited to: \\
- Violence or gore \\
- Adult or sexual content\\
- Hate symbols or discriminatory imagery\\
- Dangerous activities or self-harm\\
- Illegal substances or activities\\
- Harassment or bullying content\\
- Misinformation or manipulated media\\

Please determine if the image contains unsafe content. To finish the task, you need to FIRST think about the reasoning process as an internal monologue and then provide the final answer. The reasoning process MUST BE enclosed within <thinking> </thinking> tags.\\

Next, respond only with a JSON object in the format \{"label": "Safe"\} if the image does not contain unsafe content or \{"label": "Unsafe"\} if it contains unsafe content. Do not provide any explanation. The final answer MUST BE put in \textbackslash boxed\{\}.
    \end{tcolorbox}
    \caption{Prompt for RL on Unsafe Bench.}
    \label{fig:rl_prompt_unsafe}
\end{figure*}

\begin{figure*}
  \begin{tcolorbox}
 You are an experienced image quality assessment expert with expertise in evaluating technical aspects of digital imagery. Your task is to evaluate whether a given image has good technical quality. \\

You will be provided the following information: \\

The image to evaluate: <|image|> \\

Please carefully analyze the image's technical quality based on the following criteria:\\
- Sharpness and clarity: Is the image clear and well-focused, or is it blurry and out of focus?\\
- Compression artifacts: Are there visible pixelation, blocky patterns, or compression-related distortions?\\
- Noise and grain: Does the image have excessive noise, grain, or visual static?\\
- Distortion: Are there any warping, stretching, or other geometric distortions?\\
- Resolution: Is the image resolution sufficient for clear viewing, or is it overly low-resolution?\\
- Overall visual integrity: Can the image content be clearly perceived without significant technical impediments?\\

Consider the image to have "good technical quality" if it has minimal to no technical issues that would impair viewing or comprehension of the content.\\

Please determine if the image has good technical quality. To finish the task, you need to FIRST think about the reasoning process as an internal monologue and then provide the final answer. The reasoning process MUST BE enclosed within <thinking> </thinking> tags.\\

Next, respond only with a JSON object in the format \{"label": "Yes"\} if the image has good technical quality or \{"label": "No"\} if it does not have good technical quality. Do not provide any explanation. The final answer MUST BE put in \textbackslash boxed\{\}.
    \end{tcolorbox}
    \caption{Prompt for RL on AGIN-Tech.}
    \label{fig:rl_prompt_agin_tech}
\end{figure*}

\begin{figure*}
  \begin{tcolorbox}
 You are an experienced visual content analyst with expertise in evaluating the logical consistency and rationality of images. Your task is to evaluate whether a given image is rational. \\

You will be provided the following information: \\

The image to evaluate: <|image|> \\

Please carefully analyze the image's rationality based on the following criteria:\\
- Physical plausibility: Do objects, people, and scenes follow realistic physical laws and proportions?\\
- Logical consistency: Do the elements in the image make sense together in the given context?\\
- Anatomical accuracy: Are human or animal bodies depicted with correct anatomy and natural poses?\\
- Spatial coherence: Are objects positioned and sized appropriately relative to each other?\\
- Contextual appropriateness: Do the elements fit logically within the scene or setting?\\
- Absence of absurdities: Are there any impossible, nonsensical, or surreal elements that defy rationality?\\

Consider the image to be "rational" if it depicts a coherent, logically consistent scene with minimal to no irrational or absurd elements.\\

Please determine if the image is rational. To finish the task, you need to FIRST think about the reasoning process as an internal monologue and then provide the final answer. The reasoning process MUST BE enclosed within <thinking> </thinking> tags.\\

Next, respond only with a JSON object in the format \{"label": "Yes"\} if the image is rational or \{"label": "No"\} if the image is not rational. Do not provide any explanation. The final answer MUST BE put in \textbackslash boxed\{\}.
    \end{tcolorbox}
    \caption{Prompt for RL on AGIN-Rat.}
    \label{fig:rl_prompt_agin_rat}
\end{figure*}

\begin{figure*}
  \begin{tcolorbox}
 You are an experienced image authentication expert with expertise in distinguishing AI-generated images from real photographs. Your task is to evaluate whether a given image looks natural. \\

You will be provided the following information: \\

The image to evaluate: <|image|> \\

Please carefully analyze the image's naturalness based on the following criteria:\\
- AI generation artifacts: Are there visible signs of AI generation such as unnatural textures, blending issues, or synthetic patterns?\\
- Photorealism: Does the image exhibit the natural characteristics of a real photograph, including authentic lighting, shadows, and depth?\\
- Detail consistency: Are fine details (skin texture, hair strands, fabric, surfaces) rendered naturally and consistently?\\
- Object and human features: Do faces, hands, objects, and their features appear natural without distortions or uncanny elements?\\
- Visual coherence: Does the overall image maintain photographic quality without obvious artificial or synthetic areas?\\

Consider the image to be "natural" if it looks as authentic as a real photograph with minimal to no visible unnatural or AI-generated characteristics.\\

Please determine if the image looks natural. To finish the task, you need to FIRST think about the reasoning process as an internal monologue and then provide the final answer. The reasoning process MUST BE enclosed within <thinking> </thinking> tags.\\

Next, respond only with a JSON object in the format \{"label": "Yes"\} if the image looks natural or \{"label": "No"\} if it does not look natural. Do not provide any explanation. The final answer MUST BE put in \textbackslash boxed\{\}.
    \end{tcolorbox}
    \caption{Prompt for RL on AGIN-Nat.}
    \label{fig:rl_prompt_agin_nat}
\end{figure*}

\begin{figure*}
  \begin{tcolorbox}
 You are an experienced visual-linguistic analyst with expertise in evaluating the alignment between textual descriptions and visual content. Your task is to evaluate whether a given text description matches a given image.\\

You will be provided the following information: \\

Text Description: \{Image Description\}\\
The image to evaluate: <|image|> \\

Please carefully analyze whether the text description accurately matches the image based on the following criteria:\\
- Content accuracy: Do the objects, people, animals, or scenes described in the text appear in the image?\\
- Detail consistency: Are the specific details mentioned in the text (colors, positions, actions, attributes) consistent with what is shown in the image?\\
- Completeness: Does the text describe the main elements visible in the image without introducing elements that are not present?\\
- Contextual alignment: Does the overall context or scene described in the text match the visual context of the image?\\

Consider the text and image to "match" if the text accurately describes the visual content of the image without significant discrepancies or contradictions.\\

Please determine if the text description matches the image. To finish the task, you need to FIRST think about the reasoning process as an internal monologue and then provide the final answer. The reasoning process MUST BE enclosed within <thinking> </thinking> tags.\\

Next, respond only with a JSON object in the format \{"label": "Yes"\} if the text matches the image or \{"label": "No"\} if the text does not match the image. Do not provide any explanation. The final answer MUST BE put in \textbackslash boxed\{\}.
    \end{tcolorbox}
    \caption{Prompt for RL on SeeTrue.}
    \label{fig:rl_prompt_seetrue}
\end{figure*}

\begin{figure*}
  \begin{tcolorbox}
You are an experienced visual-linguistic analyst with expertise in evaluating the alignment between textual descriptions and visual content. Your task is to evaluate whether a given text description matches a given image.\\

You will be provided the following information: \\

Text Description: \{Image Description\}\\
The class of the image: \{Classification\}\\
The image to evaluate: <|image|> \\

Please carefully analyze whether the text description accurately matches the image based on the following criteria:\\
- Content accuracy: Do the objects, people, animals, or scenes described in the text appear in the image?\\
- Detail consistency: Are the specific details mentioned in the text (colors, positions, actions, attributes) consistent with what is shown in the image?\\
- Completeness: Does the text describe the main elements visible in the image without introducing elements that are not present?\\
- Contextual alignment: Does the overall context or scene described in the text match the visual context of the image?\\

Consider the text and image to "match" if the text accurately describes the visual content of the image without significant discrepancies or contradictions.\\

Please determine if the text description matches the image. To finish the task, you need to FIRST think about the reasoning process as an internal monologue and then provide the final answer. The reasoning process MUST BE enclosed within <thinking> </thinking> tags.\\

Next, respond only with a JSON object in the format \{"label": "Yes"\} if the text matches the image or \{"label": "No"\} if the text does not match the image. Do not provide any explanation. The final answer MUST BE put in \textbackslash boxed\{\}.
    \end{tcolorbox}
    \caption{Prompt for RL on Image Reward.}
    \label{fig:rl_prompt_image_reward}
\end{figure*}

\end{document}